# Improved Twitter Sentiment Analysis in Influencer Identification Using Naïve Bayes and Custom Language Model

By: Angela Lin

## Abstract

In the last couple decades, social network services like Twitter have generated large volumes of data about users and their interests, providing meaningful business intelligence so organizations can better understand and engage their customers. All businesses want to know who is promoting their products, who is complaining about them, and how are these opinions bringing or diminishing value to a company. Companies want to be able to identify their high-value customers and quantify the value each user brings. Many businesses use social media metrics to calculate the user contribution score, which enables them to quantify the value that influential users bring on social media, so the businesses can offer them more differentiated services. However, the score calculation can be refined to provide a better illustration of a user's contribution. Using Microsoft Azure as a case study, we conducted Twitter sentiment analysis to develop a machine learning classification model that identifies tweet contents and sentiments most illustrative of positive-value user contribution. Using data mining and AI-powered cognitive tools, we analyzed factors of social influence and specifically, promotional language in the developer community. Our predictive model was a combination of a traditional supervised machine learning algorithm and a custom-developed natural language model for identifying promotional tweets, that identifies a product-specific promotion on Twitter with a 90% accuracy rate.

## Background

Microsoft Azure is a set of cloud services that enable developers to build, deploy, and manage applications through Microsoft's global network of datacenters [3]. Azure's cloud computing model seeks to emphasize scalability, agility, flexibility as differentiating features of their cloud platform from their main competitor, Amazon Web Services [11]. Currently, Microsoft Azure (Microsoft's cloud business) uses social media metrics to calculate the "user contribution score". This enables Microsoft to quantify the value that influential users bring to its cloud business on social media, so Microsoft can offer them more differentiated services. However, the score calculation needs to be refined to provide a better illustration of a user's contribution.

Some sentiment analysis has already been done on the tweets, classifying whether the tweets about Azure are positive or negative, but many sentiment analysis tools are not the most accurate for the purposes of identifying a user's "contribution" [4,5]. A tweet may be classified as positive not because it is necessarily expressing a positive remark about Azure, but rather because it includes keywords such as "good" or "awesome" with an exclamation mark at the end. Similarly, a tweet may be classified as neutral because it expresses negative sentiment toward Azure's competitor, AWS, but positive sentiment toward Azure. Generic sentiment analysis tools are only capable of identifying polarity thus overlook such nuances that need to be considered in developing an accurate way of measuring how a user's opinion (over social media) towards a product brings or diminishes value to the company. The average accuracy of all sentiment analysis tools is only 54% [5]. As a result, just a sentiment label is not very representative of the tweet content nor a good measurement of a user's "contribution" over social media.

Thus, to conduct more targeted sentiment analysis, we first define a promotional tweet as any tweet that can bring positive value to a Microsoft by: 1) spreading positive ideas about Azure, 2) influencing users to use it more frequently or 3) helping users learn more about the functionalities of the product and its services.

We thus use this framework to outline the most defining attributes of an Azure promotional tweet, when building a custom language model specific to the context of Azure and ultimately training our

classification model to predict "promotional" or "non-promotional."

## Literature Review:

Most Twitter sentiment analysis studies have relied on a knowledge base approach or machine learning approach; in our research, we adopted a hybrid approach. Like the study conducted by Saif et. Al (2012), we also took the approach of adding semantics as additional features in the training set [20], using extracted entities such as "Azure Features." Saif et. al showed that the Recall and F score both increased with this approach. More importantly, this approach allows us to capture the nuances of promotional tweets. Another sentiment classification study on Twitter data by Barbosa and Feng (2010) proposes the use of syntax features like hashtags, retweets, links, punctuation, and exclamation marks in conjunction with polarity and POS of words for sentiment analysis. The study utilizes manually labelled tweets for both tuning and testing. We extend this approach by incorporating similar Twitter syntax features our data attributes and follow the same methodology in training and testing our data. Damon (2004) performed sentiment analysis on feedback data from Global Support Services through extensive feature analysis and feature selection and demonstrated that abstract linguistic analysis features contribute to the classifier accuracy. In this paper, we also perform extensive feature analysis before training our language model.

There are many existing classification algorithms that are standards for text classification, most notably Naïve Bayes. After careful analysis, this is the algorithm chosen to classify whether a tweet is considered an Azure promotion or not due to its versatility, speed of training, and high accuracy rate in information retrieval. Because Naïve Bayes does not need a large training set to accurately predict its classes, it is also one of the most efficient classifier models [12,19].

## Methods and Protocols:

**Technical Overview:**

We used two main applications: Waikato Environment for Knowledge Analysis (WEKA) for our supervised machine learning model and Microsoft's Language Understanding Intelligence Service (LUIS) for the custom-developed natural language model. LUIS uses the power of machine learning to extract meaning from natural language input and allows the user to define a custom language model. LUIS "takes a user utterance and extracts intents and entities" that correspond to activities in the application's logic developed from training input [10]. The utterance is the textual input from the user that LUIS interprets; in this case, the tweets. The intent represents actions the user wants to perform; in this case, identifying whether a tweet is "promotional" or not. The entity "represents an instance of a class of object that is relevant to a user's intent," which translates to the defining attributes of a promotional tweet [10]. LUIS uses state-of-art natural language processing techniques and models, which are built into its natural language understanding abilities. Because of the tool's learning efficiency, usability, and high-level intelligence capabilities, LUIS was the optimal tool to use to capture the complexities of a promotion sentiment. Given the various ways different groups of users express themselves on social media, especially in writing microtexts like tweets [7], LUIS's customizability enabled us to illustrate the nuances of natural language. LUIS also has a programmatic API key created automatically that allows the user to publish the model to the endpoint in the format of a URL, which we implemented in our code automating the data attribute analysis. The final output is a predicted intent (either promotional or non-promotional, in our case) with an intent score, a numeric value between 0 and 1 (1 being the highest) that represents the model's prediction confidence level.

WEKA [13] is a Java-based open-source data mining tool that we used to develop a classification model to predict unseen data after using a training data set. WEKA was used for various tasks including our data pre-processing and comparative analysis between other algorithms. After training the classifier to our scenario, we evaluated, refined and eventually applied the classifier to different scenarios.

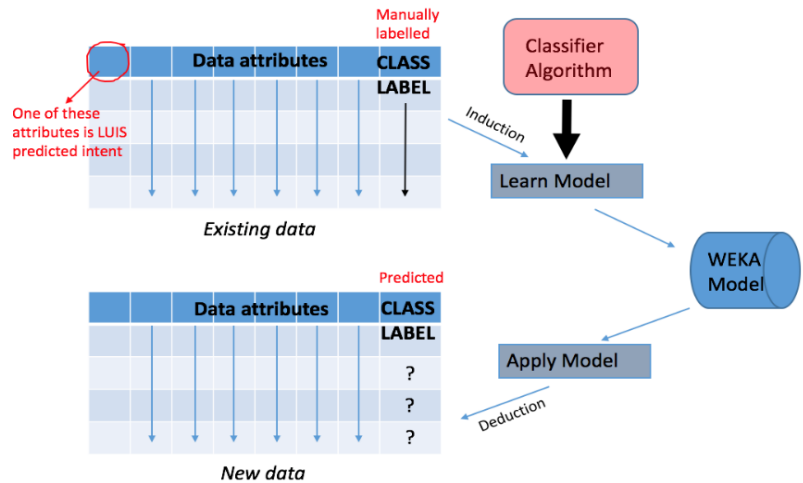

**Figure 1**: WEKA Machine Learning Classification Illustration

**Procedure**

1. Using a Twitter API call [17], we streamed all the tweets containing the keywords of "Azure", "AWS", and "Cloud" from the top influencers in the developer community based on Twitter followings.
2. We organized the raw data in a SQL database for further analysis. Sampling 100 random tweets from 10,000 total, we first manually labelled the tweets promotional or non-promotional with a brief explanation.
3. All manually labelled promotion data were cross-checked by a total of three other employees on the Microsoft Data Science team and Azure Social Media team. In-person meetings and investigation during the data mining stage led to the team consensus about the logic and standard of what qualifies as a "promotion" tweet. [See Mentors/Affiliates in Section IX]
4. We analyzed each tweet, identifying and recording observations of attributes and features that constitute a promotional or non-promotional tweet based on the standard we developed within the Microsoft Data Science team for the specific purpose of this project. We compiled lists of reoccurring words that appeared in tweets, what we called "keywords," that were relevant to the "promotion" sentiment [Appendix B].
5. Once a preliminary list of important attributes was developed, we tested how accurate those attributes were in determining an Azure promotion sentiment by measuring the independent strength of each attribute in determining a promotional tweet. All attributes identified at this stage were manually parsed from raw Twitter data, and eventually narrowed down to 8-10 most important attributes.
6. Using the logic behind the linguistic attributes identified, we trained LUIS to predict a promotion intent. Most of the logic was translated in the form of LUIS's "entities" and "intents" [10]. Keywords that were identified in preliminary attribute research were stored as "phrase lists" in LUIS.
7. Once the LUIS model was mature, we implemented the intent prediction and score as two more attributes into the final supervised machine learning model we would train and develop. [See model logic in Section V to understanding this reasoning.]
8. Using 36 of the 100 random tweets as the first training data with all the attributes manually labeled per tweet, we started studying the proper algorithm we would use to classify the tweets as promotional or non-promotional. Top contestants were Naive Bayes, decision tree, linear regression. The first training data set did not include the LUIS results as attributes.
9. We wrote a program to produce random samples of tweets, cleanse the data and automatically identify the attributes of each tweet in the format needed for the machine learning classifier to read. Data attributes were automatically written them into a CSV file to be uploaded to WEKA.
10. We trained the newest model that includes the LUIS results, using our program to conduct data attribute analysis on a random sample of 38 tweets as the model input.
11. We then applied it to new data the model had not yet encountered. Running the program again, we collected our necessary data inputs (attributes) to test our WEKA model. In determining our model's accuracy, we compared the model's predicted results to our manually determined result – whether, given the data attributes, the tweet should be promotional or not. It is important to note that all tweets being tested were manually labelled before WEKA produced its predicted results, to avoid bias.
12. Using the program we wrote, we continued the refining process by sampling new tweets, testing newly defined attributes, modifying our LUIS model until the optimal model with the optimal algorithm could identify promotion tweets about Azure with a 90% accuracy rate. With the most recent model, we conducted 35 rounds of testing, two of which are documented in this paper and analyzed the incorrectly predicted tweets.

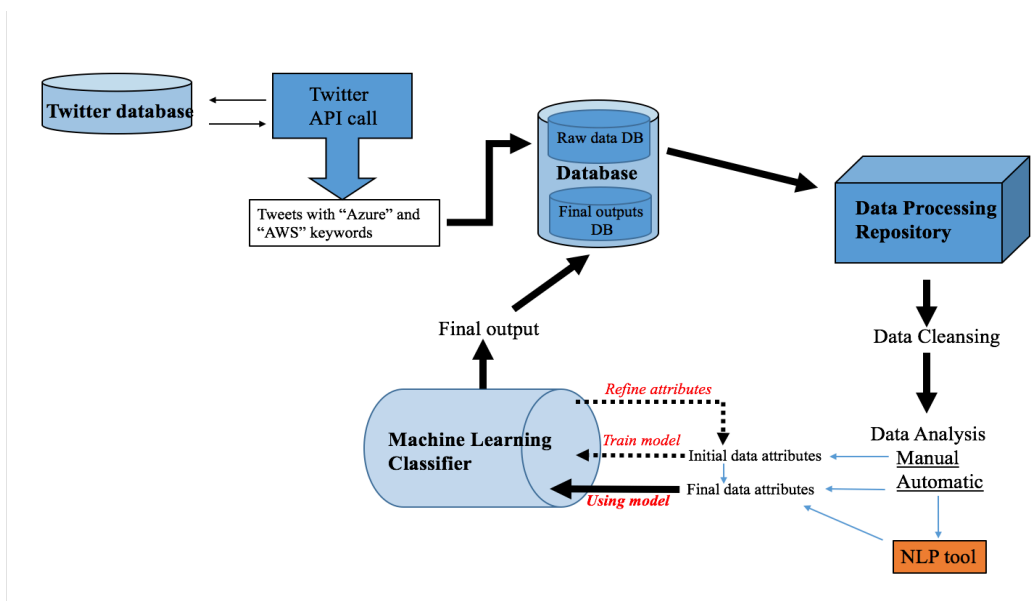

**Figure 2:** System Architecture

## Data Processing and Analysis:

After our initial raw data analysis of "Azure"-keyword-related tweets, we discovered 72% of the tweets were classified as "neutral," indicating that most tweets by developers did not carry extreme emotion – regardless of the tweet's contributive value. Moreover, between 48% and 60% of the total tweets classified as "neutral" would be labelled as "promotional" based on our standards (estimated with a 90% confidence, using a sample of 150 tweets), showing how influential tweets may not always be associated with a positive sentiment. This indicated there are more defining attributes than a positive polarity measurement that ultimately determined a promotional sentiment.

What we did notice was the use of the same kind of language or keywords in all the tweets that we labelled as promotion. Thus, many of our data attributes relied on keyword based knowledge extraction, one of the most precise ways of conducting tweet classification and sentiment analysis [4]. Using actual tweet examples from our experiment, below shows the logic behind parsing each tweet and analyzing the attributes that make a tweet "promotional."

**"Promotion" Sentiment Analysis Examples:**

Tweet Example 1: **PROMOTION**

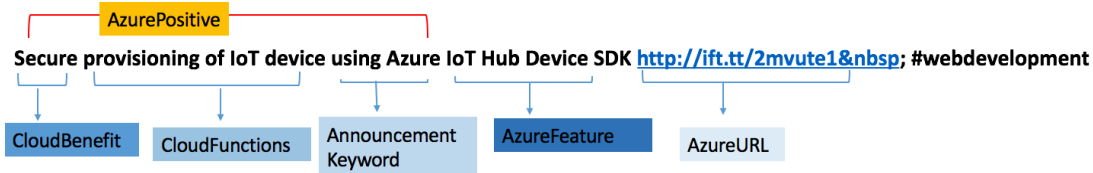

**Figure 3:** Parsing Promotion Tweet Example 1

Tweet Example 2: **PROMOTION**

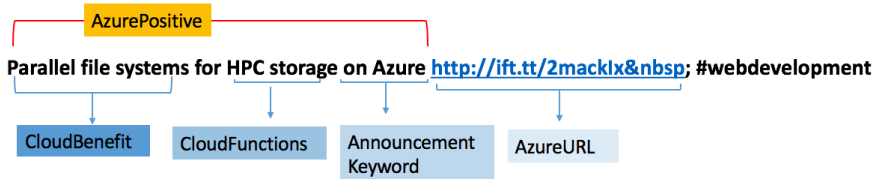

**Figure 4:** Parsing Promotion Tweet Example 2

Tweet Example 3: **PROMOTION**

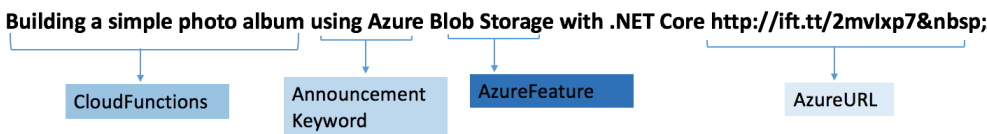

**Figure 5:** Parsing Promotion Tweet Example 3

Tweet Example 4: **PROMOTION**

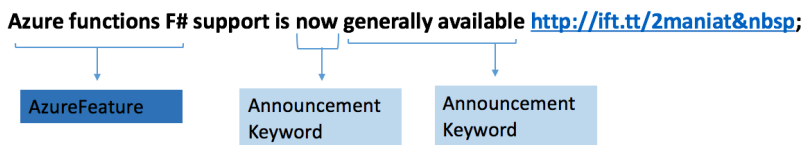

**Figure 6:** Parsing Promotion Tweet Example 4

Example 1 and 2 show clear benefits that Azure offers that is specific to its services, and thus are marked accordingly with the "AzurePositive" label. The "AzurePositive" label is a parent entity that usually will include an "CloudBenefit" keyword and some "CloudFunction," as seen above. All include Azure-specific URL's. In both examples, the benefit or clear advantage conveyed is specified with phrases like "using Azure" and "on Azure" to indicate it is the service that Azure is providing that is beneficial or advantageous.

However, Example 3 is still considered a promotional Tweet because although not a specific benefit is mentioned related to Azure, the Tweet still brings attention to an important feature or capability (in this case, building a photo album) that Azure Blog Storage [6] can do. An Azure-specific URL is included in the Tweet that provides a link to a Microsoft blog with further information. Example 4 is a clear example of a promotional tweet that is mainly an announcement, calling attention to a new feature – which also carry a lot of impact.

Tweet Example 5: **NON-PROMOTIONAL**

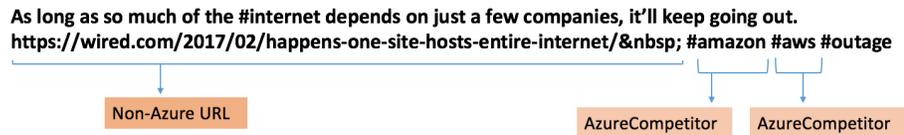

**Figure 7:** Parsing Non-Promotion Tweet Example 5

Tweet Example 6: **NON-PROMOTIONAL**

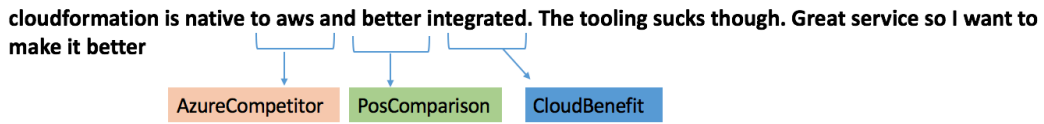

**Figure 8:** Parsing Non-Promotion Tweet Example 6

Example 5 is classified as non-promotional because it is non-specific to Azure. It discusses a broader cloud computing/internet trend that makes it too generic to include any useful or impactful information about Azure – not to mention it also mentions Azure's competitors. Example 6 is non-promotional for Azure because it's a promotion Tweet for Azure's competitor, AWS – indicated by "AzureCompetitor" being in the same sentence as an "CloudBenefit", which shows the user is showing positive sentiment towards Azure's competitor, praising a benefit that should otherwise be Azure's. Moreover, the absence of any Azure-related keyword makes it non-specific, and thus, irrelevant to Azure.

We list simplified examples to show how we parse tweets that mention both Azure and its competitor.

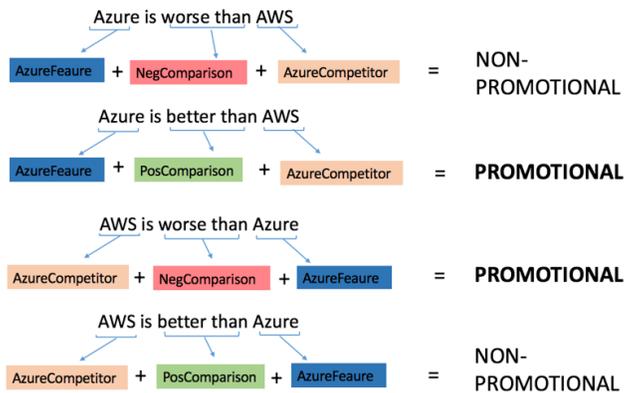

**Figure 9:** Parsing Azure vs. AWS

Below we have listed the identifiers of a promotion sentiment that we compiled from the twitter parsing stage, from a natural language processing standpoint. Many of them contain keyword libraries or phrase lists used to help the language model identify and learn instances of the identifiers.

| **AzureFeature** | Specific features/capabilities that Azure has (usually nouns specific to Azure services); identifies each any Azure-related words like "Azure" itself or "Microsoft" to ensure the tweet is specific to Azure. |
|---|---|
| **CloudFunction** | Capabilities that cloud platform services have *in general*; used to determine in the context of the identified cloud capability or feature is being discussed about generically or in specific reference to Azure |
| **AnnouncementKeyword** | Keywords that indicate when tweets are calling attention to some new feature, or added benefits; also ensures that some cloud computing function or added benefit mentioned in the tweet is specific to Azure (why "on Azure," "with Azure," or "using Azure" phrases are included in this keyword library) |
| **CloudBenefit** | Identifies "benefit" keywords, which are usually action verbs like "manage", "network", "enhance" or adjectives like "faster" and "serverless" to really illustrate the impact a cloud service is making and the value its capabilities can bring to the user<br>• *Want to know if an identified benefit from a cloud service is the benefit that Azure is bringing*<br>• *LUIS model trained to associate the presence of a "benefit" tag with an "AzureCompetitor" tag as a "non-promotional" tweet whereas the combination of "benefit" and "AzureFeature" is much more likely a "promotional" tweet* |
| **AzurePositive** | Composite entity (composed of CloudFunction, AzureFeature, CloudBenefit, Announcement) that is helpful to identify any time someone talks about something good that Azure is doing and the impact it's making at large, or how Azure has helped people in specific tasks in cloud computing; composite because usually this is a phrase, not just a word that will include words from the other entities<br>• *Triggered when one of the three non-composite entities are identified in conjunction with each other*<br>• *Includes "Announcement" for "using/on Azure," phrases* |
| **PosComparison** | Phrase list that identifies phrases like "better than" to identify how Azure is being compared to its competitor |
| **NegComparison** | Same logic as PosComparison but with phrases like "worse than" |
| **EqualComparison** | Similar logic to PosComparison and NegComparison identifiers; helpful to know when a user identifies Azure services to be equal to its counterparts |
| **AzureCompetitor** | Helpful to know when a tweet mentions another cloud service like AWS that is Azure's competitor so further analysis can be done according to logic in previous section |
| **QuantifiedImpact** | Identifies quantified impact that Azure or Microsoft brings (either to individuals or organizations) like speed, percentages, money |

**Figure 10:** List of Model Attributes

**Relevant non-natural language identifiers:**

**URLs:**
- The presence of a URL indicates that the tweet contains a link to another website or picture, thus showing there is more information to be told or explained. Given the 140-character limit of a tweet, a link to an article or blog is a more effective way of sharing information than the brief catchphrase in a tweet body. The content of the tweet body is often just a way to further promote information in the article or blog shared.

**Azure URL:**
- If the URL is Microsoft or Azure-specific, the URL is either a link to a Microsoft or Azure website or an article or blog specifically about Microsoft or Azure (See link in [6] for example.)
- A URL to a Microsoft website or Azure blog is always a positive identifier as Microsoft would never express negative sentiment toward its own product. Most examples have been the Azure blog website that explain a new feature or discuss troubleshooting techniques. These are all tweets we want to see because they are a) highlighting Azure's hallmarks or key traits b) showing the user how they can improve what they're currently working on with Azure solutions (thus increasing frequency of using the product and probably satisfaction level) and c) bringing the user to the Microsoft website, which in it of itself is a marketing tool.

**Punctuation:**
- A tweet containing a colon indicates the user is most likely announcing something and an exclamation usually indicates the intent to bring attention to a certain message. In shorter tweets where the tweet body is just a title of an article and the link, the use of a colon is an important indicator that the user is or "announcing" or calling attention to the content in the link.

**Insights about Twitter Promotion Data:**

We discovered that twitter promotions are subtler, more nuanced and often unintentionally done. The person tweeting may have just wanted to share a new Azure feature with fellow developers or an article about a Microsoft blog that's helpful in accomplishing some task using Azure services. Without knowing it, these users *are* helping promote Azure in a way that is influencing other developers to use Azure more frequently or more knowledgeably, which would ultimately enable a more positive user experience.

At first glance, many of the tweets seemed to be more informational than promotional (neutral sentiment scores, short tweets with no particular description), but that was because those tweets tended to be the title of an article or blog post on the Microsoft/Azure website that was included in the tweet. Oftentimes, the promotion wasn't the tweet body itself; it was the link that led to another site that would then discuss the positive aspects/benefits of Azure [20]. Thus, Twitter potentially becomes an avenue to get more page views and trafficking on Microsoft's website, which is the most effective marketing tool.

Most tweets that are not classified as Azure promotion tweets have been classified that way not because they express negative sentiments toward Azure but because they are non-specific towards Azure. In fact, there were few instances when users actually complained or criticized Azure services/Microsoft. On average, only about 8% of the non-promotional classified tweets were considered actual negative sentiments/complaints about Azure. *Positive or negative sentiment does not directly correlate to positive or negative contributions that the user is making.*

# **Our Model Approach**

The goal is to identify all the relevant linguistic and non-linguistic attributes of a "promotion" sentiment. We can use LUIS to do the first and WEKA to do the latter. The reason why we used both LUIS and WEKA was because given the particular scenario, both tools have their limitations. First we analyze the text classifier we used, Naïve Bayes, and how we exploit its framework to make our model more robust.

**Applying Naïve Bayes to Azure business scenario:**

Naïve Bayes is a probabilistic model that uses the Bayes theorem to solve the maximum posterior probability (MAP) of a class label (c) given its attributes set (d):

$$C_{MAP} = \underset{c \in C}{argmax}\ P(c|d) \qquad (1)$$

$$= \underset{c \in C}{argmax}\ \frac{P(d|c)P(c)}{P(d)} \qquad (2)$$

Dropping the denominator and showing the attribute set as each attribute feature,

$$\underset{c \in C}{argmax}\ P(d_1, d_2, ..., d_n\ |\ c)P(c) \qquad (3)$$

In Naïve Bayes, we are trying to determine the most likely class, in this case, "Azure promotion" or none, given the conditional probabilities ($P(x|c)$) of the attributes of each tweet. In our case, $P(x|c)$ measures the frequency that an attribute is present in a tweet that is classified as 'promotional.' Our classifier model can be represented mathematically,

$$\underset{c_j \in C}{argmax}\ P(c_j) \prod P(x_i\ |\ c_j) \qquad (4)$$

which translates to:

$$P(d_1, d_2, ..., d_n\ |\ c) = P(x_1\ |\ c) \cdot P(x_2\ |\ c) \cdot P(x_3\ |\ c) \cdot\ .... \cdot P(x_n\ |\ c_{j=}) \qquad (5)$$

Even though the independence assumption of Naïve Bayes does not always hold true for textual data, as there will always be dependence between certain attributes [12], it provides a framework to think about our problem: The goal is to maximize each $P(x|c)$ to maximize the chance that the model's prediction of a certain class *really is* that certain class. In other words, our selection of attributes need to be informative enough so that we are not feeding the model noisy data – attributes that have little correlation with an actual Azure promotion tweet. Below is the investigation of our main initial list of attributes and how we determined the effectiveness of each attribute.

**Data Attribute Evaluation:**

Below are graphs that measure the effect each individual attribute from the attribute list attained by Twitter parsing (without NLP tools) has on the identification of an Azure promotion (See Step 6 in Methodology). The red indicates the number of tweets classified as promotional and blue indicates the number of tweets not classified as promotional. There were a total of 36 in this sample.

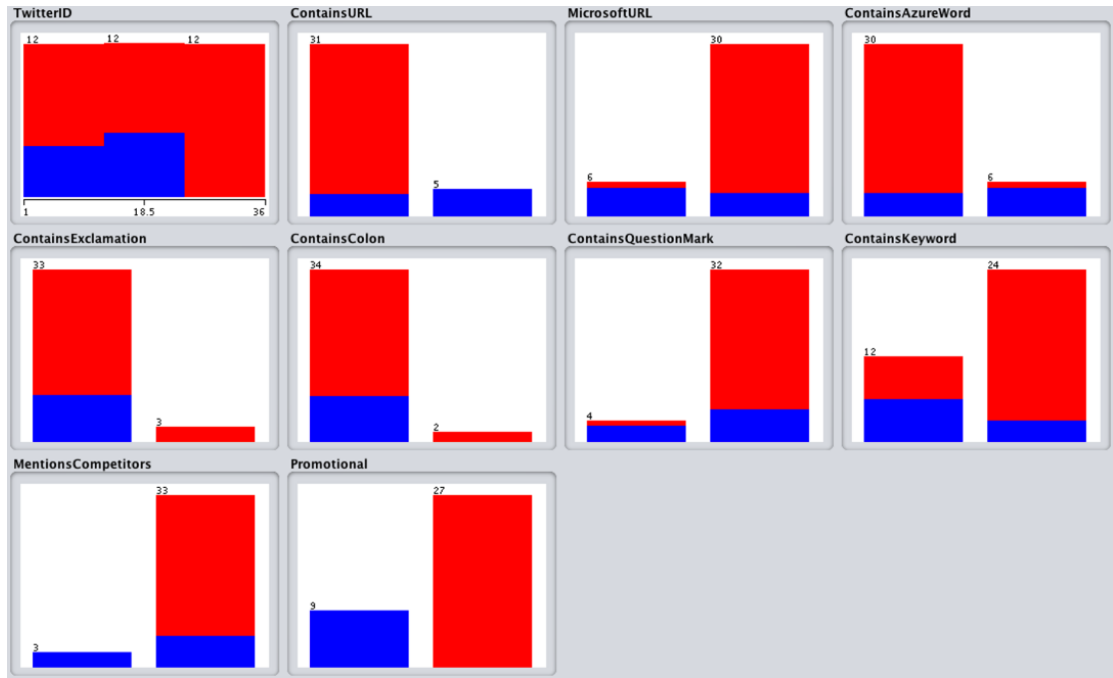

**Figure 11:** Data produced by WEKA visualization tool

ContainsURL and MicrosoftURL: **Indicative of Promotion**
- All tweets without URLs were classified as non-promotional (doesn't mean this relationship is automatically causal, just an observation from this sample), and 87% of the Tweets that did contain a URL were classified as promotional, mostly because those URLs were a Microsoft or Azure specific one.
- Out of the tweets that had Microsoft-specific URLs, 86% of them were classified as promotional.

Contains AzureWord: **Highly indicative of Promotion**
- 87% of tweets classified as promotional mentioned Azure or some Azure-related word (proof for why a tweet must be Azure-specific to be considered promotional). Similarly, all but one of the tweets that did not have an Azure-specific word were classified as non-promotional, showing that on average, generic Tweets without mentioning "Azure" are highly likely to be non-promotional - because they are usually some trend discussion or informational tweet about a different topic.

Contains Keyword (an "announcement" keyword): **Highly indicative of Promotion**
- Out of the Tweets that contained a keyword, 83% were classified as promotional, showing that if a tweet has an announcement keyword, it has a high chance of being a promotional one. At this point, half of the tweets that didn't contain an announcement keyword were still classified as promotional. When re-analyzing these tweets, we discovered this was because our announcement keyword library could still be expanded to include other examples of announcement keywords.

MentionsCompetitors**: Highly indicative of non-promotional**
- Out of the tweets that didn't mention a competitor of Azure, 85% of them were classified as promotional while all of the tweets that did mention an Azure competitor were all non-promotional. If an Azure competitor appears in a tweet, it is highly probable that the tweet is non-promotional.

Punctuation (colon, question, exclamation):
- The absence of a punctuation mark doesn't affect the chances of a Tweet being promotional or not (as many of the tweets that are classified as promotion have neutral sentiments anyway). In other words, a tweet without a punctuation mark does not automatically indicate it is non-promotional. However, the presence of a punctuation mark can make a difference. All of the Tweets that had a colon or exclamation mark in the sample set were classified as promotion tweets—a correlational not causational relationship.

**Combining Statistical and Natural Language Approach:**

While the first several rounds of Twitter attribute parsing and attribute refinement eventually led to the discovery of highly relevant data attributes, we discovered the overall prediction accuracy of the model improved with the addition of a couple other features. The first was an addition of another keyword library that specifically identified a *benefit*, an addition made after realizing how effective it worked as an "object" in the LUIS model. The addition of this new feature reduced the instances that the model incorrectly classified a tweet as promotional because it contained an Azure-specific word but not necessarily a specific benefit Azure was bringing.

The last two were the predicted intents themselves from the LUIS model. Ultimately, LUIS will develop a model that gives a predicted intent (promotional or not) and a score that indicates its confidence of its predicted intent. We discovered incorporating that intent and score as two of the attributes for the WEKA model can help validate its accuracy. Both tools have individual limitations but using them together can fully address the linguistic and non-linguistic aspects of a promotional tweet about Azure, and both the independent and dependent nature of the data attributes. LUIS conducts natural language analysis of the tweets themselves whereas WEKA just analyzes the statistical significance of the tweet attributes. Moreover, it is important to note that the logic behind the promotion sentiment analysis done from natural language parsing (examples in Data Processing section) can only be applied to the LUIS model because it relies on dependence among the class attributes: certain combinations among the "identifiers." However, WEKA is still a crucial tool because we must understand the frequency of the attributes and the overall statistical significance of these attributes, quantifying how much the attributes affect the determination of a Tweet being promotional/non-promotional. Moreover, while LUIS can learn speech and text patterns, it is limited in evaluating the content of the URLs and whether the URL in the tweet is a Microsoft/Azure-specific URL, another important attribute.

Implementing the LUIS predicted intent and score as two attributes makes the model more robust - in case there is a scenario with unusual attribute results from the binary attributes, we can get more reliability from a second analysis by the LUIS model for extra reinforcement. In other words: trying to make $\prod P(\mathbf{x}_i | \mathbf{c}_i)$ as high as possible, the total conditional probability of all the independent attributes, by ensuring $P(\mathbf{x}_{\text{LUIS predicted intent}} | \mathbf{c}_{\text{promotion}})$ is always high. Looking at the training set [see WEKA Model data], we can see that given a LUIS predicted intent of promotional and a score of at least 0.5, the probability the final prediction of a tweet being promotional is 1. Similarly, given a LUIS prediction of the opposite, the final prediction will likely be non-promotional.

When on the fence, the LUIS results can tip the final prediction in the right direction. Furthermore, adding the LUIS predicted intent and score increased the model accuracy rate from 86% (determined from model will attributes except for the two LUIS predicted results) to 97%, based on WEKA's stratified cross-validation of the training data set. (97% is not the final model accuracy.)

The LUIS model is trained to a pretty high degree of accuracy given its NLP tools; however, the LUIS cannot effectively conduct URL analysis. LUIS can identify keywords that appear in the URL but does not recognize it as a web link, nor does it have the ability to open it and analyze the content. That's where we depend on our program to conduct URL analysis and WEKA to take into account the relevant URL information as data attributes.

### Final Model attributes for WEKA: (all are binary, yes or no, except score)

1. **Is there a URL?**
2. **Is the URL a Microsoft or Azure-specific URL?**
3. **Does the tweet contain the word "Azure"** (or an Azure-related word)?
   - If the Tweet does not at least contain "Azure," it can't be considered promotional because it won't be specific to Azure.
4. **Is there an "announcement" keyword in the tweet?**
   - Same as LUIS "announcement keyword" entity; see earlier section for logic
5. **Is there a question mark?**
   - Inquiring about something, could be rhetorical
6. **Is there a colon?**
   - Colons are used to introduce something; evidence that user may be announcing or introducing an important feature
7. **Is there an exclamation mark?**
   - Indicates excitement or emphasis, a strong emotion that adds weight to whatever message is being sent
8. **Is there an "Azure-specific benefit" keyword?**
   - Same as LUIS "announcement keyword" entity; see below for LUIS model logic
9. **Does the tweet mention a competitor of Azure?**
   - Very important indicator in understanding what is being said about Azure against its competitors especially in the same tweet. More logic in LUIS model in comparing Azure against its competitors.
10. **What is the final intent prediction that LUIS returns?** (promotion = yes; non-promotional=no)
    - See above for logic on implementing LUIS into WEKA
11. **What is that intent score?** (on a scale from 0-1)
    - See above for logic on implementing LUIS into WEKA

**LUIS model:**

Utterances:
- Actual tweet examples that we collected in data mining stage

Intents:
- Either a promotion or non-promotion (indicated by LUIS as "none")

Entities:
- All the natural language processing identifiers listed in the table above; both the entities of the LUIS model and the attributes of the final WEKA model contain the same keyword lists (just with different names). The presence of a keyword is important enough that it should be a final attribute, but is also a fundamental entity in helping LUIS determine the final prediction.

Phrase lists:
- To help LUIS learn quicker, we compiled phrase lists for nearly all the entities.

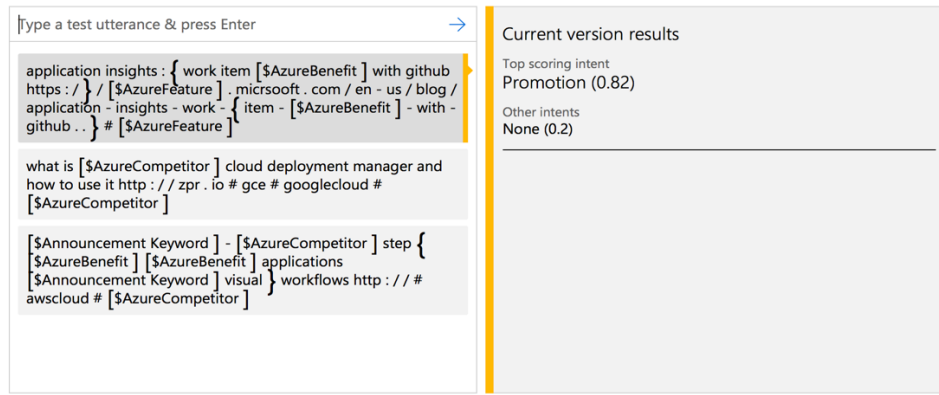

**Figure 12:** Example of how LUIS analyzes utterances *(AzureBenefit label name changed to "CloudBenefit", content is same)*

**Data Processing Program:**

Our Azure Twitter Promotion Program was written to automate the data processing stage in order to analyze large volumes of tweets.

Program functionalities
1. Produces random samples of tweets from original raw data and keeps track of tweets that have been fetched
2. Data cleansing: (link to code: Data Cleansing and Processing)
    a. Removed tweets that were not in English (this model only considers tweets in English)
    b. Removed all the "#" signs as the LUIS API doesn't recognize this symbol, thus leading inaccurate sentiment analysis
    c. Removed all "&nsbp" as part of the URL's which rendered most as invalid URL's, retrieved original URL
    d. Used Regex to identify all original URL sources for URLs that were shorthand (URL's that end in "t.co")
3. Data Attribute Analysis based on final WEKA model attributes
    a. Regex to read and understand the data better for keywords analysis too
    b. All keywords stored in libraries to easily identify if such keywords exist
    c. URL Analysis: To identify whether a URL was "Microsoft-specific," we looked for the words "Microsoft" or "Azure" in the URL and we fetched the title of the URL article link to see if the words "Microsoft" or "Azure" appeared there. We also ensured there was no mention of Azure's competitor (AWS, Amazon etc.).
    d. Programmed automatic call to LUIS API so tweets could automatically receive intent prediction and score
4. Writes all data attributes results into a CSV file to be uploaded into WEKA (CSV Writer)

## Data Results:

**TRAINING SET:** Random sample of 38 tweets, labeled and analyzed by our program: (See Appendix A for the actual tweets the data attributes correspond to.)

| ContainsURL | MicrosoftURL | ContainsAzureWord | ContainsExclamation | ContainsColon | ContainsQuestionMark | ContainsKeyword | MentionsCompetitor | ContainsBenefit | Intent | IntentScore | Promotional |
|---|---|---|---|---|---|---|---|---|---|---|---|
| y | y | y | n | n | n | y | y | y | y | 0.99 | y |
| n | n | y | n | n | n | n | n | y | y | 0.61 | y |
| y | y | y | y | n | n | n | n | n | y | 0.5 | y |
| y | y | y | n | n | n | y | n | y | y | 0.66 | y |
| y | y | y | n | n | n | y | n | n | y | 0.94 | y |
| y | y | y | n | n | n | y | n | y | y | 0.97 | y |
| y | y | y | y | n | n | y | n | n | y | 0.99 | y |
| y | y | y | n | y | n | y | n | n | y | 0.98 | y |
| y | y | y | n | n | n | y | n | y | y | 0.99 | y |
| y | n | y | y | n | n | y | n | y | y | 0.99 | y |
| y | y | y | n | n | n | y | n | n | y | 0.96 | y |
| y | y | y | n | n | n | y | n | y | y | 1 | y |
| y | y | n | n | n | n | y | n | n | y | 0.92 | y |
| y | y | y | n | n | n | y | n | n | y | 0.87 | y |
| n | n | y | y | n | n | y | n | n | y | 0.94 | y |
| y | n | n | n | n | n | y | y | y | n | 0.74 | n |
| y | n | n | n | n | n | y | y | n | n | 0.62 | n |
| y | n | n | n | n | n | y | y | y | n | 0.75 | n |
| y | n | n | n | n | n | y | y | n | n | 0.73 | n |
| y | y | n | n | y | n | n | n | y | y | 0.82 | y |
| y | y | y | n | n | y | y | n | n | y | 0.83 | y |
| n | n | n | n | n | n | n | n | n | n | 0.85 | n |
| n | n | n | y | n | n | n | n | n | n | 0.87 | n |
| y | y | y | n | n | n | y | n | n | y | 0.82 | y |
| y | y | y | n | n | n | n | n | n | y | 0.87 | y |
| y | n | y | n | n | y | n | n | y | n | 0.58 | n |
| y | y | n | n | n | n | y | n | y | y | 0.95 | y |
| n | n | y | n | n | n | n | n | y | n | 0.74 | n |
| y | n | y | y | n | n | y | n | y | y | 0.96 | y |
| n | n | y | n | n | n | n | n | y | y | 0.86 | y |
| n | n | y | n | n | n | y | n | y | y | 0.57 | y |
| n | n | n | n | n | n | n | y | n | n | 0.98 | n |
| y | n | y | n | n | n | y | y | n | n | 0.66 | n |
| y | n | n | n | n | n | n | y | n | n | 0.97 | n |
| y | y | y | n | n | n | y | n | y | y | 0.87 | y |
| n | n | n | n | n | n | n | n | n | n | 0.98 | n |
| n | n | y | n | n | n | n | y | n | y | 0.67 | y |
| n | n | y | n | n | n | n | n | n | n | 0.74 | n |

**Table 1:** Training Data Set

This section shows the training and testing of the final model we developed that includes the LUIS predicted intents. Using cross-validation on the training set, we developed a model that would accurately predict a Twitter promotion tweet about 90% of the time. We also conducted 35 rounds of testing the model on never-seen-before data, sampling random batches of 10 tweets each time. In total, there were 34 incorrectly predicted instances out of 350 total. Below we show the data from the first two rounds and analyze one of the incorrectly predicted instances that is representative of why other instances were incorrectly predicted by our model.

**Table 2:** Stratified cross-validation

| | | |
|---|---|---|
| Correctly Classified Instances | 37 | **97.3684 %** |
| Incorrectly Classified Instances | 1 | 2.6316 % |
| Kappa statistic | 0.9426 | |
| Mean absolute error | 0.0467 | |
| Root mean squared error | 0.1584 | |
| Relative absolute error | 10.2507 % | |
| Root relative squared error | 33.1673 % | |
| Total Number of Instances | 38 | |

**Table 3:** Stratified cross-validation

| | TP Rate | FP Rate | Precision | Recall | F-Measure | MCC | ROC Area | PRC Area | Class |
|---|---|---|---|---|---|---|---|---|---|
| | 0.960 | 0.000 | 1.000 | 0.960 | 0.980 | 0.944 | 0.997 | 0.998 | y |
| | 1.000 | 0.040 | 0.929 | 1.000 | 0.963 | 0.944 | 0.997 | 0.995 | n |
| Weighted Avg. | 0.974 | 0.014 | 0.976 | 0.974 | 0.974 | 0.944 | 0.997 | 0.997 | |

**Model Validation Using Testing Data Set:**

Table 1

| ContainsURL | MicrosoftURL | ContainsAzureWord | ContainsExclamation | ContainsColon | ContainsQuestionMark | ContainsKeyword | MentionsCompetitor | ContainsBenefit | Intent | IntentScore | WEKA predict | Error predicti | ACTUAL |
|---|---|---|---|---|---|---|---|---|---|---|---|---|---|
| y | y | y | n | n | y | y | n | n | y | 0.8777028 | Promotion | 0.999 | Promotion |
| y | n | y | n | n | y | n | y | n | n | 0.8309562 | None | 0.997 | None |
| y | y | y | n | n | n | n | n | y | y | 0.596603 | Promotion | 0.999 | Promotion |
| y | y | y | n | n | n | y | n | y | y | 0.8243235 | Promotion | 1 | Promotion |
| y | n | y | n | n | y | y | n | y | y | 0.913607 | Promotion | 0.897 | Promotion |
| y | y | y | n | n | y | n | n | n | y | 0.8756058 | Promotion | 1 | Promotion |
| y | y | y | n | n | y | n | n | y | y | 0.7027494 | Promotion | 1 | Promotion |
| y | y | y | n | n | y | n | y | y | y | 0.8328837 | Promotion | 1 | Promotion |
| y | n | y | n | n | n | n | n | n | n | 0.5263253 | None | 0.952 | None |
| y | y | y | n | n | y | n | n | y | y | 0.8715698 | Promotion | 1 | Promotion |

**Table 4:** Testing Data Set Round 1

Table 1

| ContainsURL | MicrosoftURL | ContainsAzureWord | ContainsExclamation | ContainsColon | ContainsQuestionMark | ContainsKeyword | MentionsCometitors | ContainsBenefit | Intent | IntentScore | Weka Predic | ACTUAL |
|---|---|---|---|---|---|---|---|---|---|---|---|---|
| True | n | y | n | y | n | n | n | n | y | 0.6634236 | Promotional | Promotional |
| True | n | n | n | y | n | y | n | n | n | 0.9565789 | None | None |
| True | n | y | n | y | y | y | n | n | y | 0.9212027 | Promotional | Promotional |
| False | n | n | y | y | n | n | n | y | n | 0.7019325 | None | Promotional |
| False | n | y | n | n | n | n | n | n | y | 0.8399622 | Promotional | Promotional |
| True | n | n | n | y | n | n | n | n | n | 0.9181368 | None | Promotional |
| True | y | y | n | y | n | n | n | n | n | 0.8733456 | Promotional | Promotional |
| True | y | y | n | y | n | y | n | n | n | 0.5381471 | Promotional | Promotional |
| True | y | y | n | y | n | y | n | n | y | 0.8366119 | Promotional | Promotional |
| True | y | y | y | y | n | n | n | n | y | 0.9042173 | Promotional | Promotional |

**Table 5:** Testing Data Set Round 2

Incorrectly predicted tweets:

As noted above, Tweet #4, a promotional tweet was incorrectly identified as non-promotional, most likely because it didn't contain an announcement keyword and because the LUIS predicted intent was non-promotional, with a relatively high intent score. The LUIS prediction attribute is most likely to sway the final WEKA prediction the most, due to its high conditional probability in the training set. As seen in the training set, there are no instances when the LUIS prediction did not match the actual result, which is advantageous because $P(x|c)$ is maximized, but also relies on the LUIS model to be extremely accurate. After referring to the LUIS breakdown of this tweet, we discovered the most likely reason LUIS did not identify the utterance as promotional was because it did not register "MS" (initials for Microsoft) as an Azure-related word or "w/" as the word "with." Thus, it could not trace the benefit of "cross-platform cloud power" as relating to an Azure benefit. This example shows that the LUIS model phrase lists needs to be continually updated based on Twitter shorthand language and other nuances in natural language to accurately identify its entities. The LUIS model also did not recognize the smiley face emoticon, which would otherwise be associated with a positive sentiment towards Azure or Microsoft in conjunction with the presence of the "w/ MS" phrase.

The other incorrectly predicted tweets followed the same logic: the majority were either predicted incorrectly due to the limitations of the keyword libraries included in our program or as a result of an incorrect LUIS predicted intent. Several were incorrectly predicted because our program was unable to process unusual character entities and URLs presented in certain formats. More research needs to make the data cleansing stage more robust.

## **Conclusions and Further Investigation:**

In this study, we created a machine learning model that conducts natural language analysis to predict promotion tweets about a cloud service with a 90% accuracy rate. As a first step, companies like Microsoft can use this model to analyze and predict promotion tweets, both at an individual level and on a larger scale among large volumes of tweets that discuss cloud services. As a second step, because the key attributes of a promotion tweet have been identified, companies can quantify users' influence or contribution on Twitter by measuring the strength of a promotion based on how many promotion attributes are present in a tweet. Data attributes can be exploited to shed more light on *what extent* they are discussing Azure in a positive light. Was it a *general* announcement about a new Azure feature or did it mention a *specific* Azure benefit? Were there keywords, hashtags or URLs linking to the Microsoft website? While the final WEKA model only produces an output of "promotion" or "non-promotional," our program indicates all the data attributes of each tweet it analyzes before the data is sent to the WEKA model. The final contribution algorithm can take into account the presence of such data attributes to produce a more meaningful influencer identification or contribution score for businesses. The binary nature of the data attributes allows them to be quantified much easier. The logic outlined in this project can help companies more accurately identify key influencers for Azure on Twitter so Microsoft can micro-target their high-value users and offer them differentiated services.

Ultimately, this twitter promotion framework facilitates a more sophisticated sentiment analysis, taking into account the specific language, jargon, and way of communication between developers to better understand how and why developers would be influenced within their social networks when discussing cloud services. The Twitter data we analyzed in this project helped us realize that a "promotion" among developers does not necessarily correlate with an extremely "positive" sentiment, but are rather influenced by more nuanced factors. It is important to adapt the social media sentiment analysis to the demographic being analyzed. The way developers communicate over Twitter reflects the unique nature of that *specific* demographic – and thus generic sentiment analysis only investigating polarity is not sufficient. The approach and methodology employed in this project can be applied to other business scenarios to refine the social media sentiment analysis and influencer identification. Moreover, the specific model can also be adapted by other businesses by changing the keyword libraries to become more specific to the product features they want to analyze and the objectives of the respective company.

**Further investigation and improvements:**

It is important to note that the fact a user is promoting Azure on Twitter with a smaller following than another user doesn't diminish the value of the promotional Tweet analysis done in this research. The model analyzes promotion from the tweet language, not from its actual social impact. A user should still be awarded for discussing Azure in a positive light regardless of his or her reach – however the reach can be a factor taken into account in the larger calculation of the user's overall contribution score.

One aspect of the model in questioning is whether the final Naïve Bayes model needs as many attributes as we have identified, especially with the LUIS model conducting most of the analysis from the language side. We included these attributes for extra reinforcement for the final Naïve Bayes model, but further experimentation can be done to test whether the prediction accuracy level is actually improved. The goal is to reduce "noisy data" while preserving the nuances of natural language. Conversely, there are other attributes that we have not considered such as the use emoticons; further research can be done to see if such would be a useful addition.

While the logic of this model will stay the same, what could be improved and added to are the keyword libraries in both the LUIS model and the keywords that the code identifies. Sometimes an important Azure feature won't be recognized by LUIS or the code as one because it was not included in the keyword library, which could lead to skewed attribute data, thus giving the wrong kind of input to WEKA. Keyword libraries need to continue to be updated (by that same token, LUIS needs to continue to be trained to recognize those phrases or keywords) so the automated process can be as accurate as possible. Overall, the model should be tested with more data to continue refining both the LUIS model and the program written.

For now, this model has only been trained to identify when there is an Azure URL, but not to evaluate the content within the URL. Code is currently written to conduct basic URL analysis, extracting the title and identifying if it is specific to Azure. To improve the model, further sentiment analysis can be done on the title in the same way it is done for the tweet body, identifying keywords. Analysis can be done on the website article in the links as well.

## Appendix A: Tweet contents for training data analysis

| Tweet corresponding to training data entry  Tweet ID | Tweet Content |
|---|---|
| 1  740516501464190000 | Microsoft releases preview of new Azure 'serverless compute' service to take on AWS Lambda http://ift.tt/1SrRnWU  #webdevelopment |
| 2  705374842510512000 | Secure Communications Between Azure Web Apps and Virtual Machines by MVP @ThomasArdal on our |
| 3  723438672746401000 | #Azure's Open Source Journey for Cloud!https://www.youtube.com/watch?v=JiYYYup1uWM … #MVPBuzz #OSS |
| 4  717440115396706000 | Using Remote Profiling with Git deployments in Azure Web Apps by MVP @AIDANJCASEY. https://blogs.msdn.microsoft.com/mvpawardprogram/2016/04/05/using-remote-profiling-with-git-deployments-in-azure-web-apps/ … #MVPBuzz |
| 5  844928205333151000 | Announcing Azure Service Fabric 5.5 and SDK 2.5 https://azure.microsoft.com/blog/announcing-azure-service-fabric-5-5-and-sdk-2-5/ … |
| 6  846785071336079000 | How Azure Security Center helps reveal a Cyberattack https://azure.microsoft.com/blog/how-azure-security-center-helps-reveal-a-cyberattack/ … |
| 7  763144337647476000 | Get started with #Azure Table storage using .NET (recently updated!) https://azure.microsoft.com/en-us/documentation/articles/storage-dotnet-how-to-use-tables/ … |
| 8  860178415047102000 | Now Generally Available: On-premises data gateway in Azure https://azure.microsoft.com/blog/on-premises-data-gateway-functionality-goes-ga-in-azure/ … |
| 9  851434023498924000 | Azure AD Connect now support Managed Service Account https://lnkd.in/gwMU472  |

| | | |
|---|---|---|
| 10 | 835041642788835000 | You should join me at this Cyber Threat Detection and Response with Azure Meetup. Check it out and RSVP! http://meetu.ps/36Zf1D  |
| 11 | 856919501178372000 | Azure management libraries for Java generally available now http://zpr.io/PwM3z  #Microsoft #Azure #Cloud |
| 12 | 855166365564633000 | How Microsoft builds massively scalable services using Azure DocumentDB http://zpr.io/PDrz7  #Microsoft #Azure #Cloud |
| 13 | 860162861229903000 | Empowering digital transformation together at Red Hat Summit http://zpr.io/PFpbT  #Microsoft #Azure #Cloud |
| 14 | 860162861229903000 | 1400 compatibility level in Azure Analysis Services http://zpr.io/PFzdf  #Microsoft #Azure #Cloud |
| 15 | 860147780152684000 | 90% of fortune 500 use MS #azure #cloud! #msbuild |
| 16 | 790589253076090000 | Oracle Bare Metal Cloud Services Are Now Available http://zpr.io/P6Qf2  |
| 17 | 792056822052032000 | What's new with Google Cloud Resource Manager, and other IAM news http://zpr.io/Pguyz   #GCE #GoogleCloud #Google |
| 18 | 804404722472026000 | New – AWS Step Functions – Build Distributed Applications Using Visual Workflows http://zpr.io/PRVjC  #AWSCloud #CloudComputing #Amazon |
| 19 | 800800050163154000 | What is Google Cloud Deployment Manager and how to use it http://zpr.io/PRPbf   #GCE #GoogleCloud #Google |
| 20 | 758303572576665000 | Application Insights: Work item integration with GitHub https://azure.microsoft.com/en-us/blog/application-insights-work-item-integration-with-github/ … #Azure #AppInsights |
| 21 | 757629118200553000 | HDinsight – How to use Spark-HBase connector? https://blogs.msdn.microsoft.com/azuredatalake/2016/07/25/hdinsight-how-to-use-spark-hbase-connector/ … #Azure |
| 22 | 757627822982762000 | *simulated |
| 23 | 749682236803452000 | see you there! #mvpbuzz |
| 24 | 772848477026017000 | How to register U-SQL Assemblies in your U-SQL Catalog https://blogs.msdn.microsoft.com/azuredatalake/2016/08/26/how-to-register-u-sql-assemblies-in-your-u-sql- |
| 25 | 783041493061758000 | Microsoft Cloud coming to France http://bit.ly/2dVEVWT  #Azure |
| 26 | 788667500338696000 | Need help to integrate #data, moving #BI to #Azure, or #dataviz reporting? Try a #BusinessIntelligence healthcheckhttp://businessintelligencehealthcheck.com  |
| 27 | 856921793646309000 | Use this little tool to assess your Mobility Management Strategy http://microsoft.postclickmarketing.com/EMS-Assessment  #EMS #Azure |
| 28 | 868267808941121000 | I've done some both in C# and Node.  I'd really like to make security easier for hosting in Azure but haven't had the time |
| 29 | 749945430327304000 | Awesome! #GrabCaster is now able to run in #Azure #ServiceFabric,full reliable,full availability and high scaling :)pic.twitter.com/PNrf7ywZdg |
| 30 | 842870928560312000 | but I love how much simple is to manage the TokenCredentials in the AD @Azure Portal that makes a lot of sense ;-) |
| 31 | 842866234664407000 | love to use MAML on @Azure and in my opinion I definitely prefer the CertificateCloudCredentials to the TokenCredentials, easier and quicker |

| 32 845731844570042000 | Thinking about dropping AWS from Kubernetes the Hard Way. Too time consuming; I want to focus on Kubernetes configuration, not GCP vs AWS. |
|---|---|
| 33 847527747966705000 | Companies want to move to the #cloud but have no idea how http://ow.ly/raCC30afUHG  via @FortuneMagazine #azure #aws #bluemix |
| 34 838385610800902000 | The embarrassing reason behind #Amazon's huge #cloud computing outage this week http://wapo.st/2mKEAHS?tid=ss_tw-bottom … via @washingtonpost #aws |
| 35 829399786570838000 | Good news. Unified Azure Information Protection client now available. https://blogs.technet.microsoft.com/enterprisemobility/2017/02/08/azure-information-protection-december-update-moves-to-general-availability/ … I like the way you can classify documents... |
| 36 784325147440275000 | The purpose of a debate is to raise issues. It is up to the listeners to decide whether to stay on-prem or go to the cloud. |
| 37 798919688637935000 | Microsoft @Azure has 38 locations, more than both AWS and Google combined. @scottgu |
| 38 866697648044220000 | Cancelled my ticket to the Azure Red Shirt tour, due to a sales meeting. Looking forward to @ScottGu at #techorama instead, though! |

## **Appendix B: Phrase Lists/Keyword Lists** (examples, full list includes hundreds):

| Announcement | Cloud Benefit | Azure Competitor | Azure Features | Cloud Functions | Eq. Comparison | Neg. Comparison | Pos Comparison |
|---|---|---|---|---|---|---|---|
| Now Available Preview Announcing Get Introducing Releasing Launch Start Provide Advanced Improve Development Develop Getting started How to How using Azure with Azure for Azure in Azure have you seen tips use get started to Azure | Manage(s) Network Secure(s) Back-up Customize(s) Build(s) Scale(s) Support(s) Recover(ies)y Create(s) Maintain(s) Develop(s) Enhance(s) Modify(ies) Integrate(s) Utilize(s) Improve(s) Protect(s) Capabilit(ies)y Organize(s) Detection Response(s) Integration Scalable Scaling Quick | Amazon AWS Google Facebook Oracle GCP Skype EC2 | Azure Microsoft MS Blob Storage Resource Manager IoT Hub Parallel file systems Traffic Manager Azure Architecture Center Azure AD Azure Active Directory AD MS Build DocumentDB DB | Deployment Migration Storage CESI Infrastructure Database Monitor Identity IoT | Equal to Same Equivalent to Or Similar to Similar Comparable Either Any The same No different Consistent with Equivalent No different than Analogous Even Identical Akin Alike | Less than Worse than Less Worse Lower Below Lesser Under The worst Sucks more Sucks Awful compared to Abominable Annoying Inferior to | Better than More than Greater than Higher than Beyond Added to Superior to Take on Trumps Above Higher quality Over The best Great compared to Amazing compared to Good compared to best |
| join<br>great<br>good<br>awesome<br>able<br>empowering<br>new<br>coming to<br>by Azure<br>with Microsoft<br>w/ Microsoft<br>w/ MS<br>welcome to<br>free | Quicker<br>Fast<br>Faster<br>Scalable<br>Severless<br>Increase(s)<br>Solution(s)<br>Impressive<br>Expand(s)<br>Distribute(s)<br>Assess(es)<br>High<br>Full<br>Reliable<br>Easy<br>Help<br>Strong<br>Integrate(s)<br>Modify<br>Utilize(s)<br>Maintain(s)<br>Recovery<br>Compatible<br>Compatibility<br>Flexible | | | | | | |

## Acknowledgements:

Mentors consulted with during data processing stage and cross checked all labelled data"
- Dr. Jim Yang, Principle Data and Applied Scientist Manager, Microsoft
- Mirror Gu, Senior Data and Applied Scientist Manager, Microsoft
- Shijing Fang, Senior Data and Applied Scientist, Microsoft
- Michael Sun, Data and Applied Scientist, Microsoft